\documentclass[letterpaper]{article} 
\usepackage{aaai24}  
\usepackage{times}  
\usepackage{helvet}  
\usepackage{courier}  
\usepackage[hyphens]{url}  
\usepackage{graphicx} 
\usepackage{natbib}  
\usepackage{caption} 
\usepackage{algorithm}
\usepackage{algorithmic}

\usepackage{newfloat}
\usepackage{listings}
\usepackage{xr}

\usepackage{amsmath}
\usepackage{amssymb}
\usepackage{threeparttable}
\usepackage{booktabs}
\usepackage{makecell}
\usepackage{multirow}
\usepackage{subfigure}
\usepackage{bbding}
\usepackage[table]{xcolor}
\DeclareCaptionStyle{ruled}{labelfont=normalfont,labelsep=colon,strut=off} 
\floatstyle{ruled}
\newfloat{listing}{tb}{lst}{}
\floatname{listing}{Listing}
\title{An Efficient Knowledge Transfer Strategy for Spiking Neural Networks \\
from Static to Event Domain}
\author{
    Xiang He\textsuperscript{\rm 1, \rm2}\equalcontrib,
    Dongcheng Zhao\textsuperscript{\rm 1}\equalcontrib,
    Yang Li\textsuperscript{\rm 1, \rm2}\equalcontrib,
    Guobin Shen\textsuperscript{\rm 1, \rm3},\\
    Qingqun Kong\textsuperscript{\rm 1, \rm2, \rm3}\thanks{Corresponding authors.},
    Yi Zeng\textsuperscript{\rm 1, \rm2, \rm3, \rm4}\footnotemark[2]
}

\affiliations{
    \textsuperscript{\rm 1}{Brain-inspired Cognitive Intelligence Lab, Institute of Automation, Chinese Academy of Sciences, Beijing, China}\\
    \textsuperscript{\rm 2}{School of Artificial Intelligence, University of Chinese Academy of Sciences, Beijing, China}\\    
    \textsuperscript{\rm 3}{School of Future Technology, University of Chinese Academy of Sciences, Beijing, China}\\
    \textsuperscript{\rm 4}{Center for Excellence in Brain Science and Intelligence Technology, Chinese Academy of Sciences, Shanghai, China}\\
    \{hexiang2021, zhaodongcheng2016, liyang2019, shenguobin2021, qingqun.kong, yi.zeng\}@ia.ac.cn

}

\begin{document}

\maketitle

\begin{abstract}
  Spiking neural networks (SNNs) are rich in spatio-temporal dynamics and are suitable for processing event-based neuromorphic data.  However, event-based datasets are usually less annotated than static datasets. This small data scale makes SNNs prone to overfitting and limits their performance. In order to improve the generalization ability of SNNs on event-based datasets, we use static images to assist SNN training on event data. In this paper, we first discuss the domain mismatch problem encountered when directly transferring networks trained on static datasets to event data. We argue that the inconsistency of feature distributions becomes a major factor hindering the effective transfer of knowledge from static images to event data. To address this problem, we propose solutions in terms of two aspects: feature distribution and training strategy. Firstly, we propose a knowledge transfer loss, which consists of domain alignment loss and spatio-temporal regularization. The domain alignment loss learns domain-invariant spatial features by reducing the marginal distribution distance between the static image and the event data. Spatio-temporal regularization provides dynamically learnable coefficients for domain alignment loss by using the output features of the event data at each time step as a regularization term. In addition, we propose a sliding training strategy, which gradually replaces static image inputs probabilistically with event data, resulting in a smoother and more stable training for the network. We validate our method on neuromorphic datasets, including N-Caltech101, CEP-DVS, and N-Omniglot. The experimental results show that our proposed method achieves better performance on all datasets compared to the current state-of-the-art methods. 
  Code is available at https://github.com/Brain-Cog-Lab/Transfer-for-DVS.
\end{abstract}

\section{Introduction}
As the third generation of neural networks, spiking neural networks (SNNs) \citep{maass1997networks} are known for their rich neurodynamic properties in the spatial-temporal domain and event-driven advantages \citep{ roy2019towards}. 
Due to the non-differentiable properties of spiking neurons, training SNNs has been a critical area of extensive academic research.
The training of SNNs is mainly divided into the following three categories: gradient backpropagation-based methods \citep{wu2018spatio, wu2019direct, zheng2021going, shen2022backpropagation, li2022neuromorphic,  deng2022temporal}, spiking time-dependent plasticity (STDP)-based methods \citep{diehl2015unsupervised, hao2020biologically, zhao2020glsnn, dong2022unsupervised}, and conversion-based methods \citep{han2020rmp, bu2021optimal, li2022efficient, liu2022spikeconverter, li2022spike}.
With these proposed algorithms, SNNs show excellent performance in various complex scenarios \citep{stagsted2020event, godet2021starflow, sun2021quantum, cheni2021reducing}. 
In particular, SNNs have shown promising results in processing neuromorphic, event-based data due to their ability to process information in the time dimension \citep{ xing2020new,  chen2020event, viale2021carsnn}.

The visual neuromorphic data mainly refers to the dataset collected by Dynamic Vision Sensor (DVS) \citep{serrano2013128}. DVS is a bio-inspired visual sensor that operates differently from conventional cameras. 
Instead of capturing images at a fixed rate, the DVS measures intensity changes at each pixel asynchronously and records the time ($t$), position ($x,y$), and polarity ($p$) of the intensity change in the form of an event stream.
DVS has been gaining popularity in various applications due to their high dynamic range, high temporal resolution, and low latency \citep{gallego2017event, zhu2018ev, stoffregen2019event, gallego2020event}.
Despite these advantages, the long and expensive shooting process is still a significant challenge for event cameras, which makes event data acquisition difficult and small in scale, thus limiting its further development. 
In contrast, static datasets are larger in scale and more accessible. Pre-trained deep neural networks can transfer well to other static datasets. 
However, applying a pre-trained model on a static dataset directly to event data often yields suboptimal results. This result highlights a sharp challenge: While static images intuitively provide rich spatial information that may benefit event data, exploiting this knowledge remains a difficult problem.
For this reason, efficiently uncovering and utilizing the knowledge in static datasets to benefit event data is important for the widespread deployment of networks for various event data applications.

In this paper, we first analyze the domain mismatch problem between networks trained on static and event datasets.
We show that the inconsistency of feature distribution is a critical barrier to the effective transfer of static image knowledge to event data. To bridge this gap, we address the challenge from two main aspects: feature distribution and training strategy.
Regarding feature distribution, we design the knowledge transfer loss function, which consists of domain alignment loss and spatio-temporal regularization to learn the temporal-spatial domain invariant features between static images and event data. 
The domain alignment loss learns and acquires domain-invariant spatial features by reducing the marginal distribution distance between static images and event data.
The spatio-temporal regularization provides dynamically adjusted coefficients for domain alignment loss to better capture temporal features in the data. 
In terms of training strategies, we propose the sliding training strategy, in which the static image inputs are gradually replaced with event data probabilistically during the training process, resulting in a smooth reduction of the role of knowledge transfer loss and a smoother learning process. Through the validation on event datasets N-Caltech101, CEP-DVS, and N-Omniglot, our method dramatically improves the performance on these datasets. 
Overall, the main contributions of this paper can be summarized as follows:
\begin{enumerate}
  \item We propose a knowledge transfer loss function that learns spatial domain-invariant features and provides dynamically learnable coefficients by regularizing event features in the time dimension. This loss function ensures that the model contains static spatial features and has a comprehensive feature representation in the temporal dimension.
  \item We propose the sliding training strategy, in which the static image inputs are gradually replaced with event data probabilistically during the training process, resulting in a smoother and more stable learning process.
  \item We conduct experiments on commonly used event datasets to verify the effectiveness of our method. The experimental results show that the proposed method outperforms the state-of-the-art methods on all datasets. 
\end{enumerate}

\section{Related Work}
In order to solve the problem of limited labeled DVS data, previous works endeavored to explore solutions such as domain adaptation, data augmentation and the development of efficient training methods.

\subsubsection{Domain Adaptation Using Static Data.}
Using static images to facilitate learning better models in the event domain is an intuitive idea.
\citet{messikommer2022bridging} use a generative event model to classify event features into content and motion features, enabling efficient matching between the latent space of events and images.
\citet{zhao2022transformer} train a convolutional transformer network for event-based classification tasks using large-scale labeled image data via a passive unsupervised domain adaptation (UDA) algorithm.
\citet{sun2022ess} introduce event-based semantic segmentation to transfer existing labeled image datasets to unlabeled events for semantic segmentation tasks.
These works are related to ours. The difference is that we exploit the spatial domain invariant features between static and event data through domain alignment loss. Further, we use coefficients dynamically adjusted at each time step to better capture the temporal properties in the data. This allows the model to contain not only static spatial features, but also an integrated feature representation of the temporal dimension. These features can provide generalized knowledge for the SNN and enhance the original SNN structure instead of pre-training a new network with more parameters.

\subsubsection{Event-Based Data Augmentation.}
Due to the limited amount of event data, directly implementing data augmentation to increase the amount of training data is a feasible strategy.
\citet{li2022neuromorphic} propose neuromorphic data augmentation to stabilize SNN training and improve generalization.
\citet{shen2022eventmix} design an augmentation strategy for event stream data, and perform the mixing of different event streams by Gaussian mixing model, while assigning labels to the mixed samples by calculating
the relative distance of event streams. Our method is orthogonal to this category of methods, i.e., these data augmentation strategies can be used together with our proposed method.

\subsubsection{SNN Efficient Training.}
Efficient training of SNNs directly is also a way to improve the generalizability of the network.
\citet{kim2021optimizing} propose Spike Activation Lift Training to help the network to deliver information across all levels.
\citet{zhan2021effective} analyze the plausibility of central kernel alignment (CKA) as a domain distance measure relative to maximum mean difference (MMD) in deep SNNs.
A number of subsequent works have contributed to the efficient training of the SNN \citep{kugele2020efficient,fang2021incorporating,deng2022temporal,zhu2022tcja, dong2023temporal, zhao2023improving}.
Nonetheless, the performance of SNN is limited by the small amount of event data.
The motivation of this paper is to solve this problem by using static data to provide generalized knowledge transfer for event data and improve the generalization of SNN.

\section{Preliminaries}
\subsubsection{Neuron Model.}
We choose the Leaky Integrate-and-Fire (LIF) neuron model \cite{dayan2005theoretical}, the most commonly used neuron model. The update of the membrane potential $\boldsymbol{u}$ can be written as following discrete form
\begin{equation}
\boldsymbol{u}^{t+1, l} = \tau \boldsymbol{u}^{t, l} + \boldsymbol{W}^l \boldsymbol{s}^{t, l-1},
\end{equation}
where $\tau$ is leaky factor and $\boldsymbol{u}^{t, l}$ denotes membrane potential of the neurons in layer $l$ at time step $t$. 
$\boldsymbol{W}^l$ and $\boldsymbol{s}^l$ represent the weight parameters of the layer $l$ and the fired spikes in layer $l$, respectively.
The membrane potential accumulates with the input until a given threshold $V_{th}$ is exceeded, then the neuron delivers a spike and the membrane potential $\boldsymbol{u}^{t, l}$ is reset to zero. The equation can be expressed as
\begin{gather}
\boldsymbol{s}^{t, l} = H\left(\boldsymbol{u}^{t, l} - V_{th}\right) \\
\boldsymbol{u}^{t+1, l} = \tau\boldsymbol{u}^{t, l} \cdot \left( 1- \boldsymbol{s}^{t, l}\right) + \boldsymbol{W}^{l} \boldsymbol{s}^{t+1, l-1},
\label{eq3}
\end{gather}
where $H$ denotes Heaviside step function. In this paper, leaky factor $\tau$ is set to 0.5 and threshold $V_{th}$ to 0.5.
\subsubsection{Processing of Neuromorphic Data.}
The Dynamic Vision Sensor (DVS) triggers an event at a specific pixel point when it detects a significant change in brightness. Formulaically, it can be expressed as
\begin{equation}
L(x, y, t) - L(x, y, t-\Delta t) \geq pC,
\end{equation}
where $x $ and $y$ denote pixel location and $\Delta t$ means the time since last triggered event at $(x, y)$. $p$ is polarity of brightness change and $C$ is a constant contrast threshold.
In this way, DVS triggers a number of events $ \varepsilon$ during a time interval in the form $\varepsilon = \{ (x_i, y_i, t_i, p_i)\}_{i=1}^N $. 
Due to the large number of events, we integrate them into frames to facilitate processing as the previous works \cite{wu2019direct, he2020comparing, fang2021incorporating, shen2022eventmix}. Specifically, the events are divided into T slices, and all events in each slice are accumulated. The $j$-th $\left(0 \leq j \leq T-1\right)$ slice event after integration, $E(j, x, y, p)$, can be defined as
\begin{gather}
E(j, x, y, p) = \sum_{j_s}^{j_e - 1} \mathbf{1}_{x, y, p}\left(x_i, y_i, p_i\right) \\
j_s=\lfloor \frac{N}{T} \rfloor \cdot j, \quad j_e=\lfloor \frac{N}{T} \rfloor \cdot (j + 1),
\end{gather}
where $\mathbf{1}_{x, y, p}\left(x_i, y_i, p_i\right)$ is an indictor function. $j_s$ and $j_e$ are the start and end index of event in $j$-th slice.

\section{Methods}
In this section, we first show the domain mismatch problem that exists for the same network trained on static and event datasets. Then, we introduce our proposed knowledge transfer loss and sliding training strategies correspondingly in terms of feature distribution and training strategy.
\begin{figure}[t]
\centering
  \includegraphics[width=1.0\linewidth]{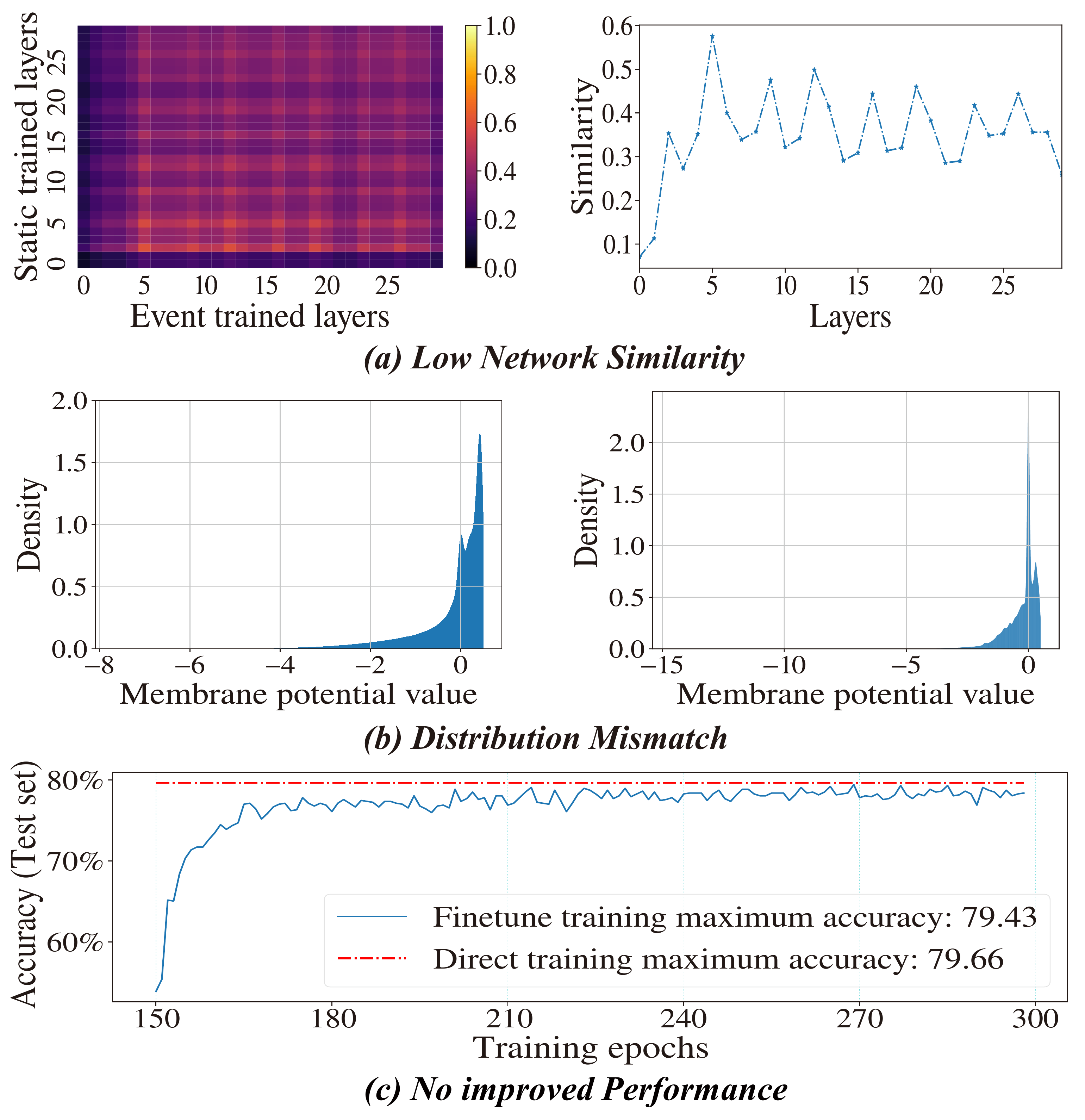}
\caption{Top: Visualization of network representation similarity. The left-left side panel shows the cross-layer heatmap, while the right side panel shows the diagonal of the cross-layer heatmap. 
Middle: Visualization of the distribution of membrane potentials.  The left and right figures show the results of the membrane potential distribution based on static data and event data training, respectively.
   Bottom: Accuracy curves when pre-trained model on static data, with fine-tuning on event data. 
   The latter half of the epochs is shown.
   }
\label{DomainMismatch}
\end{figure}

\subsection{Domain Mismatch}
Compared to static datasets, the scale of event datasets is relatively small, which makes the training more challenging. An intuitive solution strategy is to pre-train on the static dataset and then fine-tune on event dataset. However, this method suffers from a critical problem, i.e., there is a significant domain mismatch between the static and event data. To demonstrate this, we train on static dataset Caltech101 \cite{fei2004learning} and its corresponding event dataset N-Caltech101 \cite{orchard2015converting} separately using the same spiking neural network structure. 
We use the central kernel alignment (CKA) method \cite{kornblith2019similarity} to measure the similarity between features and compute CKA heatmap based on 4096 samples following \cite{nguyen2020wide, li2023uncovering}. 
Moreover, we select LIF neurons of SNN's first feature layer for membrane potential visualization. The results are shown in Fig. \ref{DomainMismatch}.

Fig. \ref{DomainMismatch}(a) shows that for the directly trained network, the features of static data are less similar to those extracted from the event dataset. In addition, the membrane potential distribution of neurons in the same layer of SNN is significantly different under different data training, as shown in Fig. \ref{DomainMismatch}(b). These results indicate that static data and event data cannot be well fused even under the same network structure. Despite the intuition that static images bring richer texture and edge information to event data, the domain difference between static and event is a hindrance.This makes the strategy of simply using static image pre-training and event fine-tuning ineffective or even counterproductive for feature extraction on event data, as shown in Fig. \ref{DomainMismatch}(c). Therefore, we need an efficient method to provide beneficial information for SNN on event data with the help of static images.

\begin{figure*}[t]
\centering
  \includegraphics[width=0.83\linewidth]{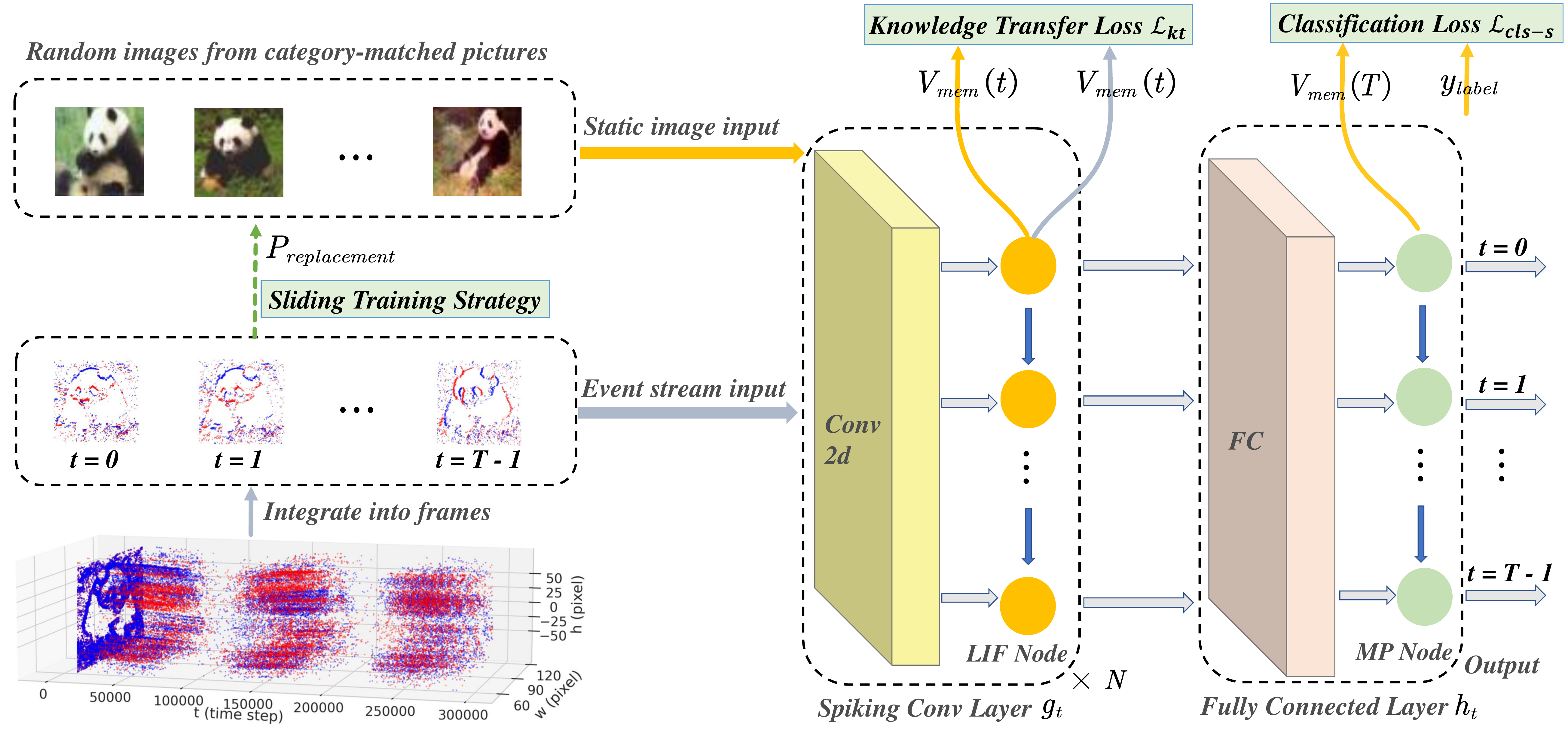}
\caption{Proposed knowledge transfer framework for spiking neural network. Static image and event data are input simultaneously and share the network weights except for the last layer. The membrane potential of the neurons in the second-last layer is used to calculate the knowledge transfer loss. 
MP node in last layer means using membrane potential output.}
\label{fig2}
\end{figure*}

\subsection{Knowledge Transfer Loss Function}
\label{s1}
The knowledge transfer loss function contains domain alignment loss and spatio-temporal regularization.

\subsubsection{Domain Alignment Loss.}
For ease of description, we first introduce some notation. We have a labeled source domain $\mathcal{D}_s=\left\{x_s^i, y_s^i\right\}_{i=1}^N$ and a small labeled target domain $\mathcal{D}_t= \left\{x_t^i, y_t^i\right\}_{i=1}^{M}$ with feature space $\mathcal{X}_s$ and $\mathcal{X}_t$ respectively. 
We aim to leverage $\mathcal{D}_s$ to assist in learning a better classifier $f_t: \mathbf{x}_t \mapsto \mathbf{y}_t$ to predict $\mathcal{D}_t$ label $\mathbf{y}_t \in \mathcal{Y}_t$.

The model for function $f$ involves a composition of two functions, i.e., $f_t=h_t \circ g_t$. Here $g_t: \mathcal{X} \rightarrow \mathcal{Z}$ represents an embedding of the input space $\mathcal{X}$ into a feature space $\mathcal{Z}$, and $h_t: \mathcal{Z} \rightarrow \mathcal{Y}$ is a function that predicts outputs from the feature space. 
We utilize the final classification head of the original model as $h_t$. This function is learned solely through supervised signal update gradients.
\textbf{Critically, we want to provide a generalization of $g_t$ which can pave the way for learning of $h_t$ to improve the generalizability of SNN.}

In this paper, the embedding function $g$ is modeled by network sharing between the source and target domains, using all layers before the last classification layer, as shown in Fig. \ref{fig2}.
At this point, the shared $g_t=g_s=g$, the optimization objective is to find the satisfied $g$ in its hypothetical space $\mathcal{G}$:
\begin{equation}
\mathop{\arg\min}\limits_{g \in \mathcal{G}} \left(d\left(g\left(X_s^a\right), g\left(X_t^a\right)\right) - d\left(g\left(X_s^c\right), g\left(X_t^a\right)\right)\right),
\label{dtarget}
\end{equation}
where $X_s^a$ and $X_t^a$ refer to the same data classes in the source and target domains while $X_s^c$ and $X_t^a$ mean the data from different classes. The $d$ is a metric for judging similarity between two domains; we choose CKA here. 
CKA is a similarity index to better measure neural network representation similarity introduced by \cite{kornblith2019similarity}. 
\begin{equation}
  \operatorname{CKA}(K, L)=\frac{\operatorname{HSIC}(K, L)}{\sqrt{\operatorname{HSIC}(K, K) \operatorname{HSIC}(L, L)}}.
\label{Eq8}
\end{equation}
where HSIC refers to Hilbert-Schmidt Independence Criterion (HSIC) \citep{gretton2005measuring} and can be computed as:
\begin{equation}
  \operatorname{HSIC}(K, L)=\frac{1}{(n-1)^2} \operatorname{tr}(K J L J),
\end{equation}
where $J$ is the centering matrix $J_n=I_n-\frac{1}{n} 11^{\mathrm{T}}$, here $I_n$ is an $n$ order unit matrix. $\operatorname{tr}$ means trace of matrix. 

To compute the CKA, we use a two-stream input paradigm: the inputs come from static image and DVS data, respectively. 
The closer the value of CKA is to 1 indicates that the two vectors are more correlated. For this reason, we subtract the CKA from 1, minimizing the loss, i.e., maximizing the correlation of the two inputs. We express the samples $\mathbf{x}_s, \mathbf{x}_t$ drawn from the whole data $\mathbf{X}_s, \mathbf{X}_t$. In this way, domain alignment loss (DAL) can be expressed as
\begin{equation}
  \small
\mathcal{L}_{d}=1 - \frac{1}{T}\sum_{t=1}^T \mathop{CKA^{\prime}}\limits_{y_i = y_j, y \in \mathcal{Y}}\left(g\left(\mathbf{x}_s^i, t\right), g\left(\mathbf{x}_t^j, t\right)\right),
\label{Eq14}
\end{equation}
where we use $g\left(\mathbf{x}_s^i, t\right)$ to indicate the value of input after shared parameter function $g$, $t$ is brought in to emphasize that here is the output of $g$ at time $t$.
Two samples $\mathbf{x}_s^i, \mathbf{x}_t^j$ are sampled from the same class, expressed by formula $y_i = y_j$.
$\mathop{CKA^{\prime}}$ represents the computation of the kernel function of the vectors followed by the computation of CKA by Eq. \ref{Eq8}.

\subsubsection{Spatio-Temporal Regularization.}
Due to the dynamic properties of event data, using only domain alignment loss for spatial feature alignment may miss important information in the temporal dimension. Spatio-temporal regularization provides dynamically learnable coefficients for the domain alignment loss, and such adaptive coefficients ensure specific weight assignments for data features at each time step. 
To prevent the model from overfitting at a certain time step, we adapt the event data classification loss at each time step (which reflects the contribution of the event frame features to the classification) as the regularization term. 
In this case, the knowledge transfer loss can be expressed as:
\begin{align}
\small
\mathcal{L}_{kt} &= 1 - \frac{1}{T}\sum_{t=1}^T \sigma(\eta_t)\mathop{CKA^{\prime}}\limits_{y_i = y_j, y \in \mathcal{Y}}\left(g\left(\mathbf{x}_s^i, t\right), g\left(\mathbf{x}_t^j, t\right)\right) \nonumber \\
&\quad + \frac{1}{T}\sum_{t=1}^T(1- \sigma(\eta_t))\ell_{cls-e},
\label{Eq15}
\end{align}
where $\eta_t$ denotes the learnable coefficient at time step $t$ and $\sigma$ represents the sigmoid function. For classification loss of event data $\ell_{cls-e}$, we choose the TET loss, which is proven to compensate the momentum loss of surrogate gradient and make SNN have better generalizability \citep{deng2022temporal}. 
$\ell_{ce}$ and $\ell_{mse}$ are the cross-entropy loss and the mean-squared loss respectively.

We add the knowledge transfer loss $\mathcal{L}_{kt}$ and classification loss of the static image $\mathcal{L}_{cls-s}$ as the total classification loss $\mathcal{L}_{all}$. 
The total training loss can be expressed as $\mathcal{L}_{all}= \lambda_{cls-s}\mathcal{L}_{cls-s} + \lambda_{kt}\mathcal{L}_{kt}$, where $\lambda_{cls-s}$ and $\lambda_{kt}$ are manually set parameters that determine the ratio of the two types of losses. 
The knowledge transfer loss not only learns domain-invariant features spatially, but also provides the network with more generalized knowledge by providing appropriate weighting coefficients temporally. This allows the model to adapt fine-grained to event data characteristics. 

\subsection{Sliding Training Strategy}
\label{s4}
The sliding training strategy aims to modulate the static image input portion of the training process so that the network gradually adapts from relying on domain-invariant features of static images and event data to fully processing event data. Specifically, during the training process, the inputs of static images are replaced by event data with probability, and this substitution probability increases with time steps until the end of the learning phase, by which time event data will replace all static images. Because the substitution process varies over time steps, as if the event data is replacing static images in a sliding time frame, we call it "sliding training". 

Separately, with $b_{i}$ denoting index of training batch, $b_{l}$ denoting total length of training batch, $e_c$ standing for current epoch and $e_m$ denoting maximum training epoch,
then the probability of making a substitution $P_{replacement}$ could be expressed by the following equation
\begin{equation}
P_{replacement} = \left(\frac{b_i + e_c * b_l}{e_s * b_l}\right) ^ 3,
\label{p-replacement}
\end{equation}
where $e_s$ is a manual settings epoch for the end of the transfer knowledge loss effects. The value of $e_s$ is usually set to $e_m$. In the early training phase, domain invariant features are dominant, providing a stable feature learning base for the model. As time advances, the proportion of event data gradually increases and the domain alignment loss gradually decreases. This gradual transition ensures the stability of the model during the learning process and avoids training instability or convergence difficulties that may result from direct or abrupt data switching.

\begin{table*}[t]
\centering
\begin{threeparttable}
  \resizebox{0.99\linewidth}{!}{
    \begin{tabular}{cccccc}
      \toprule
      \textbf{Dataset} &\textbf{Category} & \textbf{Methods} & \textbf{Architecture} & \textbf{T} & \textbf{Accuracy}\\
      \midrule
      \multirow{7}{*}{\text { N-Caltech101 }} &\multirow{2}{*}{\text { Data augmentation}} & \text { NDA} \citep{li2022neuromorphic}& \text { VGGSNN} & 10 & 78.2 \\
      & & \text {EventMixer} \cite{shen2022eventmix} & \text{ResNet-18} & 10 & 79.5\\ 
      \cmidrule{2-6}
      & \multirow{4}{*}{\text{Efficient training}} & \text { TET} \cite{deng2022temporal} & \text { VGGSNN } & 10 & $79.27 \pm 0.80 ^*$ \\ 
      & & \text{TJCA-TET} \citep{zhu2022tcja} & \text{CombinedSNN} & 14 & 82.5\\
      & & \text{TKS} \citep{dong2023temporal} & \text{VGGSNN} & 10 & 84.1\\
      & & \text{ETC} \cite{zhao2023improving} & \text{VGGSNN} & 10 & $85.53 \pm 0.09$\\
      \cmidrule{2-6}
      \addlinespace[-0.6ex]
      & \cellcolor{gray!10}{\text { Domain adaptation }}& \cellcolor{gray!10} \text { Knowledge-Transfer}  \text {(Ours) } & \cellcolor{gray!10}\text { VGGSNN } & \cellcolor{gray!10}10 & \cellcolor{gray!10}$ 93.18 \pm 0.38 \; \mathbf{(93.45)}$ \\
      \addlinespace[-0.4ex]
      \midrule
      \multirow{2}{*}{\text { CEP-DVS }} &\text { Efficient training } & \text{ TET} \citep{deng2022temporal} & \text { ResNet-18 } & 10 & $25.05 \pm 0.66 \; (25.70)^*$ \\
      \cmidrule{2-6}
      \addlinespace[-0.6ex]
      & \cellcolor{gray!10}{\text { Domain adaptation }}&\cellcolor{gray!10} \text { Knowledge-Transfer}  \text {(Ours) } &\cellcolor{gray!10} \text { ResNet-18 } &\cellcolor{gray!10} 10 & \cellcolor{gray!10} $30.05 \pm 0.50 \; \mathbf{(30.50)}$\\
      \addlinespace[-0.4ex]
      \midrule
      \multirow{3}{*}{\text { N-Omniglot }} &\multirow{2}{*}{\text { Efficient training}} & \text { plain} \citep{li2022n} & \text { SCNN } & 12 & 60.0 \\
      & &\text { plain} \citep{li2022n} & \text { SCNN } & 12 & $63.00 \pm 0.32 \; (63.44)^*$ \\
      \cmidrule{2-6}
      \addlinespace[-0.6ex]
      &\cellcolor{gray!10}\text { Domain adaptation } & \cellcolor{gray!10}\text { Knowledge-Transfer}  \text {(Ours) } & \cellcolor{gray!10}\text { SCNN } & \cellcolor{gray!10}12 & \cellcolor{gray!10} $63.60 \pm 0.46 \; \mathbf{(64.09)}$  \\
      \addlinespace[-0.4ex]
      \bottomrule
  \end{tabular}
  }
\end{threeparttable}
\caption{Experimental results compared with existing works. The results are mean and standard deviation after taking three different seeds. The best accuracy is shown in parentheses. The symbol (*) denotes our implementation.}
\label{sota}
\end{table*}

\section{Experiments}
We conduct experiments on mainstream event datasets:  N-Caltech101 \cite{orchard2015converting} and N-Omniglot to evaluate the effectiveness of the proposed method.
For another commonly used event dataset, CIFAR10-DVS \cite{li2017cifar10}, since it is 10000 samples taken from 60,000 static images from the training and test sets together, it cannot be ensured that the event data in the manually delineated test set does not overlap with the static images when using the static images to assist training. To avoid this implicit data leakage, we choose the image-event paired CEP-DVS \cite{deng2021learning} dataset as an alternative.
\subsection{Experimental Settings}
We integrate all the event data into frames and then resize to 48x48 for N-Caltech101 and CEP-DVS datasets, and for N-Omniglot dataset, it is resized to 28x28.
In terms of network structure, for a fair comparison, we choose VGGSNN (64C3-128C3-AP2-AP2-256C3-256C3-AP2-512C3-512C3-AP2-512C3-512C3-AP2-FC) model with step 10 for N-Caltech101,
Spiking-ResNet18 with step 6 for CEP-DVS, and SCNN (15C5-AP2-40C5-AP2-FC-FC) with step 12 for N-Omniglot.
For the input encoding strategy, we use direct coding for static images and 
convert the static image to HSV (Hue, Saturation, Value) color space to minimize the mismatch between the two types of input data. To adapt the dual-channel characteristics of the event data,
i.e., positive and negative polarity, we replicate the value channel and then duplicate the static image in equal time-step.
All experiments are implemented based on the BrainCog framework \citep{zeng2023braincog}.

\subsection{Comparison with the State-of-the-Art}
We first evaluate the proposed method on the N-Caltech101 dataset with VGGSNN network and compare the proposed method with NDA \citep{li2022neuromorphic}, EventMix \cite{shen2022eventmix}, TET \citep{deng2022temporal}, TJCA-TET \citep{zhu2022tcja}, TKS \citep{dong2023temporal} and ETC \citep{zhao2023improving}. The results are presented in Tab. \ref{sota}.
The experimental results demonstrate that the proposed method can achieve state-of-the-art performance compared with existing methods. 
In particular, with the proposed method, the VGGSNN network can achieve 93.45\% accuracy on the N-Caltech101 dataset. The significant performance improvement validates the effectiveness of knowledge transfer.

As for CEP-DVS and N-Omniglot datasets, there are fewer available results. 
We re-conducted the baseline experiments on these two datasets and compared them with our proposed method.
Experimental results show that our proposed method improves accuracy over the original method.
For the N-Omniglot dataset, the improvement of accuracy from knowledge transfer is not as significant as the other two datasets, this is because it is a few-shot dataset with only 20 available static images in each class, so the improvement from knowledge transfer is limited.

\subsection{Ablation Study} 
In order to verify the effectiveness of the proposed method, in the subsequent ablation experiments, we take the direct training method TET \cite{deng2022temporal} as our baseline.

\begin{table}[t]
\centering
\begin{threeparttable}
  \resizebox{0.79\linewidth}{!}{
  \begin{tabular}{ccc}
      \toprule
      Network & Methods & Accuracy\\
      \midrule
      \multicolumn{3}{c}{\textbf{N-Caltech101}}\\
      \midrule
      \multirow{4}{*}{VGGSNN} & baseline & 79.66\%\\
      & KTL w/o DAL \& STR & 84.14\%\\
      & KTL w/ DAL & 89.31\%\\
      & KTL w/ DAL \& STR & \textbf{92.64\%}\\
      \midrule
      \multicolumn{3}{c}{\textbf{CEP-DVS}}\\
      \midrule
      \multirow{4}{*}{ResNet-18} & baseline & 25.70\%\\
      & KTL w/o DAL \& STR & 27.55\%\\
      & KTL w/ DAL & 29.95\%\\
      & KTL w/ DAL \& STR & \textbf{30.50\%}\\
      \bottomrule
  \end{tabular}}
\end{threeparttable}
\caption{Ablation experiments of knowledge transfer loss on different datasets. KTL refers to knowledge transfer loss.}
\label{Ablation}
\end{table}

\begin{figure}[t]
\centering
\subfigure[Ablation experiments with the domain alignment loss and spatio-temporal regularization.]{
  \includegraphics[width=0.83\linewidth]{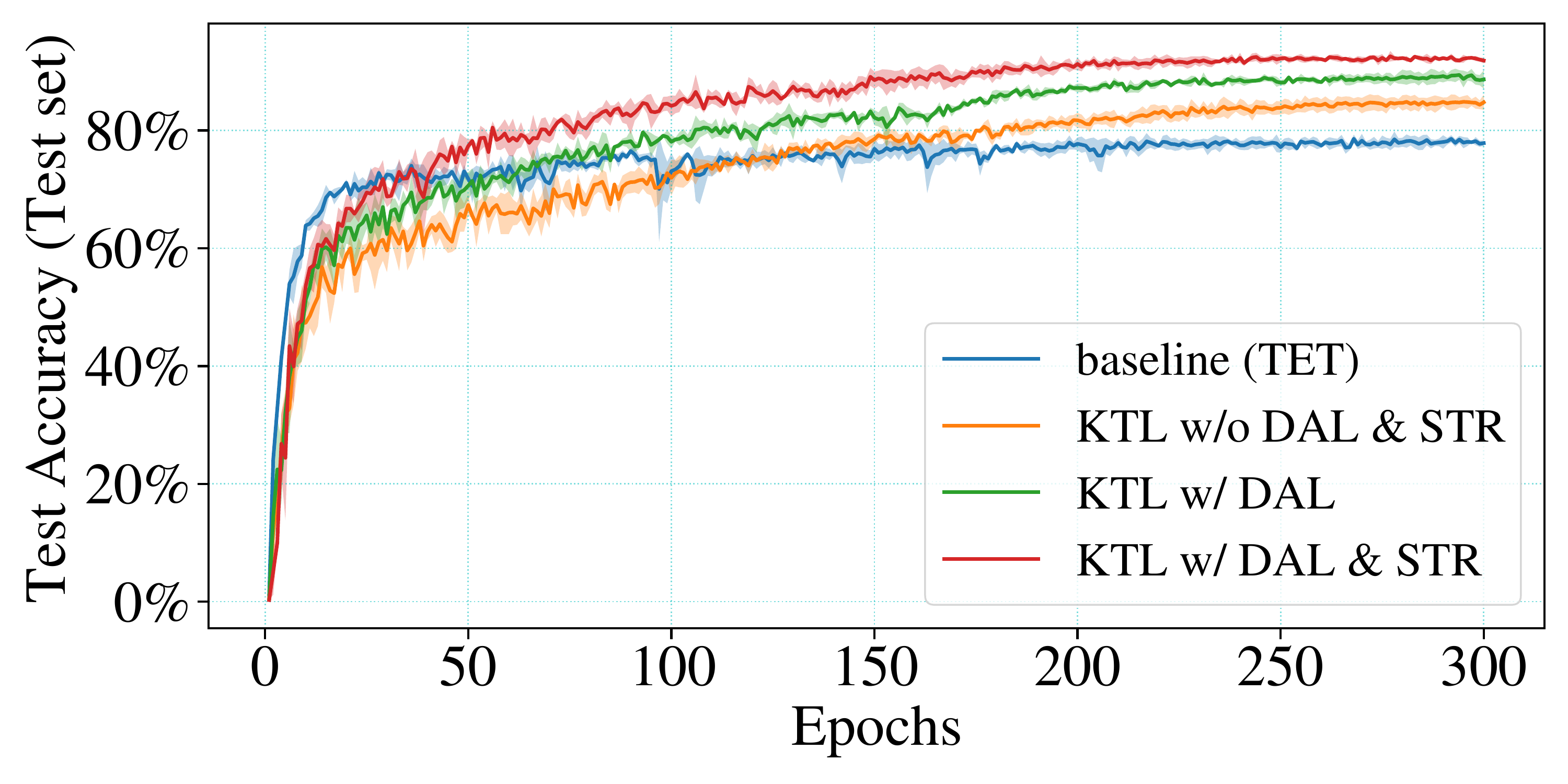}
  \label{fig:3a}
}
\subfigure[Impact of learnable coefficients on performance. Accuracy is shown at the end of each line.]{
  \includegraphics[width=0.83\linewidth]{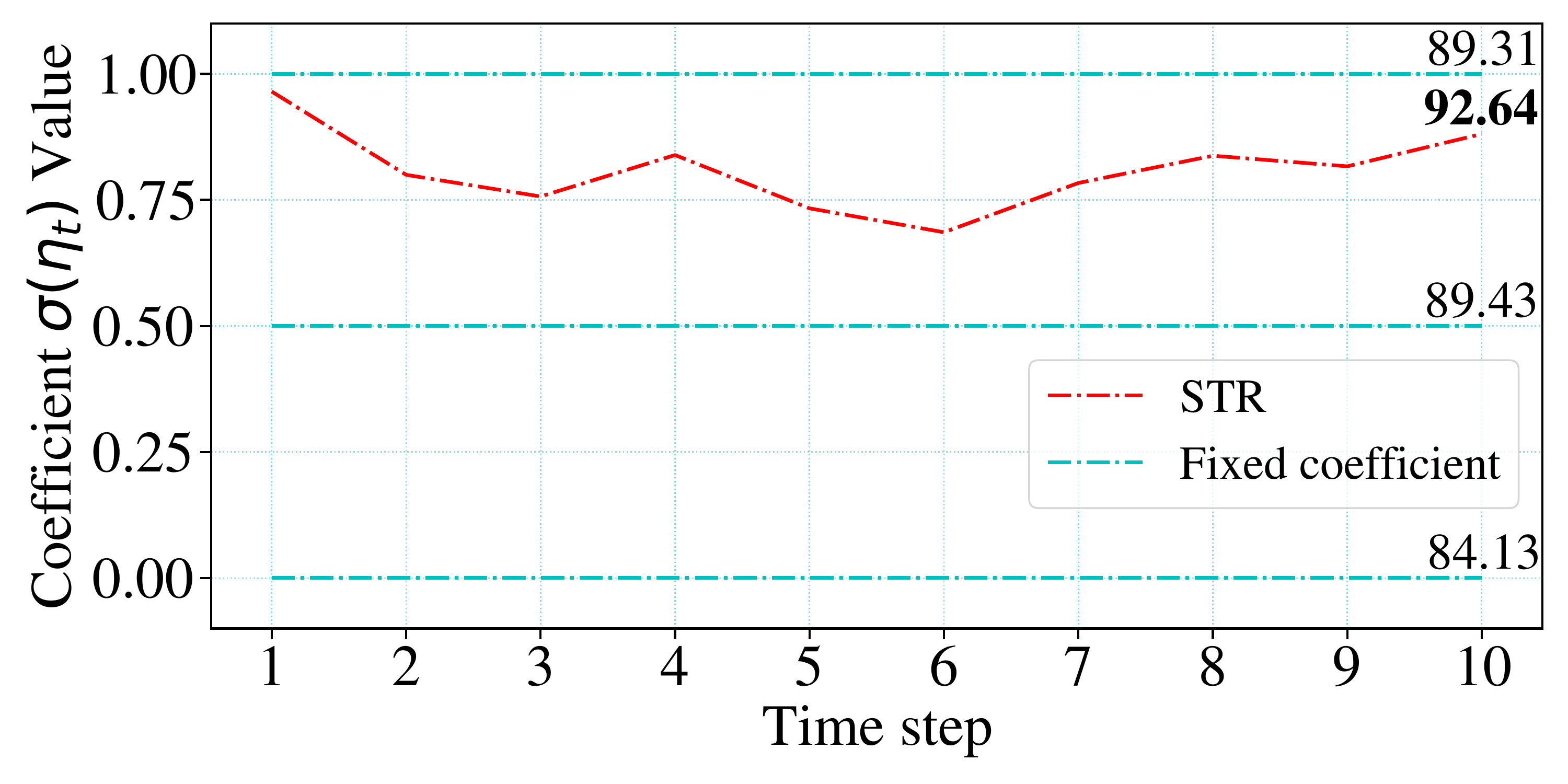}
  \label{fig:3b}
}
\caption{Performance of baseline and knowledge transfer loss methods on the N-Caltech101 dataset.}
\label{fig3}
\end{figure}

\begin{table}[t]
\centering
\begin{threeparttable}
  \resizebox{1.0\linewidth}{!}{
    \begin{tabular}{cccc}
      \toprule
      Network & Dataset &Methods & Accuracy\\
      \midrule
      \multirow{2}{*}{VGGSNN} &\multirow{2}{*}{N-Caltech101}& w/o sliding training &83.56\%\\
      & & w/ sliding training & \textbf{92.64\%}\\
      \midrule
      \multirow{2}{*}{ResNet18} &\multirow{2}{*}{CEP-DVS}& w/o sliding training &23.70\%\\
      & & w/ sliding training & \textbf{30.50\%}\\
      \bottomrule
  \end{tabular}
  }
\end{threeparttable}
\caption{Ablation experimental results for sliding training.}
\label{sliding_training}
\end{table}
\subsubsection{Knowledge Transfer Loss.} 
To verify the validity of the domain alignment loss (DAL) and the spatio-temporal regularization (STR) term in the knowledge transfer function, we conduct experiments on N-Caltech101 dataset with VGGSNN.
As shown in Fig. \ref{fig:3a}, the baseline, i.e., the TET method, has overfitted at about 100 epochs earlier. 
Compared to the baseline method, even without employing the knowledge transfer loss in our method, merely using sliding training strategy can achieve certain performance improvement.
As it gets better though with the domain alignment loss and spatio-temporal regularization to provide better generalization of the model. 
In Fig. \ref{fig:3a}, the red line is always at the top in the later training step, indicating that the best results can be achieved with these two terms.

To verify the effect of the spatio-temporal regularization, we also plot the adaptive learning coefficients of the VGGSNN at each time step under the N-Caltech101 dataset.
As shown in Fig. \ref{fig:3b}, our dynamically adjusted coefficients are superior to the coefficients that are set to be fixed at each time step, which suggests that spatio-temporal regularization to provide dynamically adjusted coefficients for the domain alignment loss is better able to capture the temporal properties in the data. In addition, the results in Fig. \ref{fig:3b} show larger coefficients at the first and last time step, which implies that the beginning and ending moment models focus more on domain-invariant spatial information.

\subsubsection{Sliding Training Strategy.} 
We conduct experiments on N-Caltech101 and CEP-DVS to verify the effectiveness of the sliding training strategy, and the results are shown in Tab. \ref{sliding_training}. The results show that sliding training leads to a more stable performance improvement. 
It is worth mentioning that in the case of without sliding training, the accuracy of our method is 23.70\%, which is slightly lower than the accuracy of the baseline method of direct training strategy, which is 25.70\%. This is due to the relatively short training epochs for CEP-DVS, which causes the model to have trouble converging in the face of sudden data switches.
Despite this, the addition of sliding training strategy solves this problem well. 

\subsubsection{Summary of Ablation Experiments.}  
We show effectiveness of each part of our proposed method with experiments of VGGSNN on N-Caltech101 dataset and the results are shown in Tab. \ref{Ablation_overview}.
The top line with no added methods is the baseline. It can be seen that without the knowledge transfer loss function, the performance of model decreases a lot. In addition, the sliding training strategy provides a guarantee for stable convergence. Combined with all the approach, our method can achieve the best performance.
\begin{table}[t]
\centering
\begin{threeparttable}
  \resizebox{0.75\linewidth}{!}{
    \begin{tabular}{cccc}
      \toprule
      DAL & STR & Sliding training & Accuracy\\
      \midrule
      -& -& -& 79.66\%\\
      \midrule
      \Checkmark& & & 82.07\%\\
      & & \Checkmark & 84.14\%\\
      \Checkmark& \Checkmark& & 83.56\%\\
      \Checkmark& & \Checkmark& 90.57\% \\
      \Checkmark& \Checkmark& \Checkmark&  \textbf{92.64\%} \\
      \bottomrule
  \end{tabular}
  }
\end{threeparttable}
\caption{Ablation experimental results overview.}
\label{Ablation_overview}
\end{table}

\begin{figure}[t]
\centering
\subfigure[Baseline, N-Caltech101]{
  \includegraphics[width=0.42\linewidth]{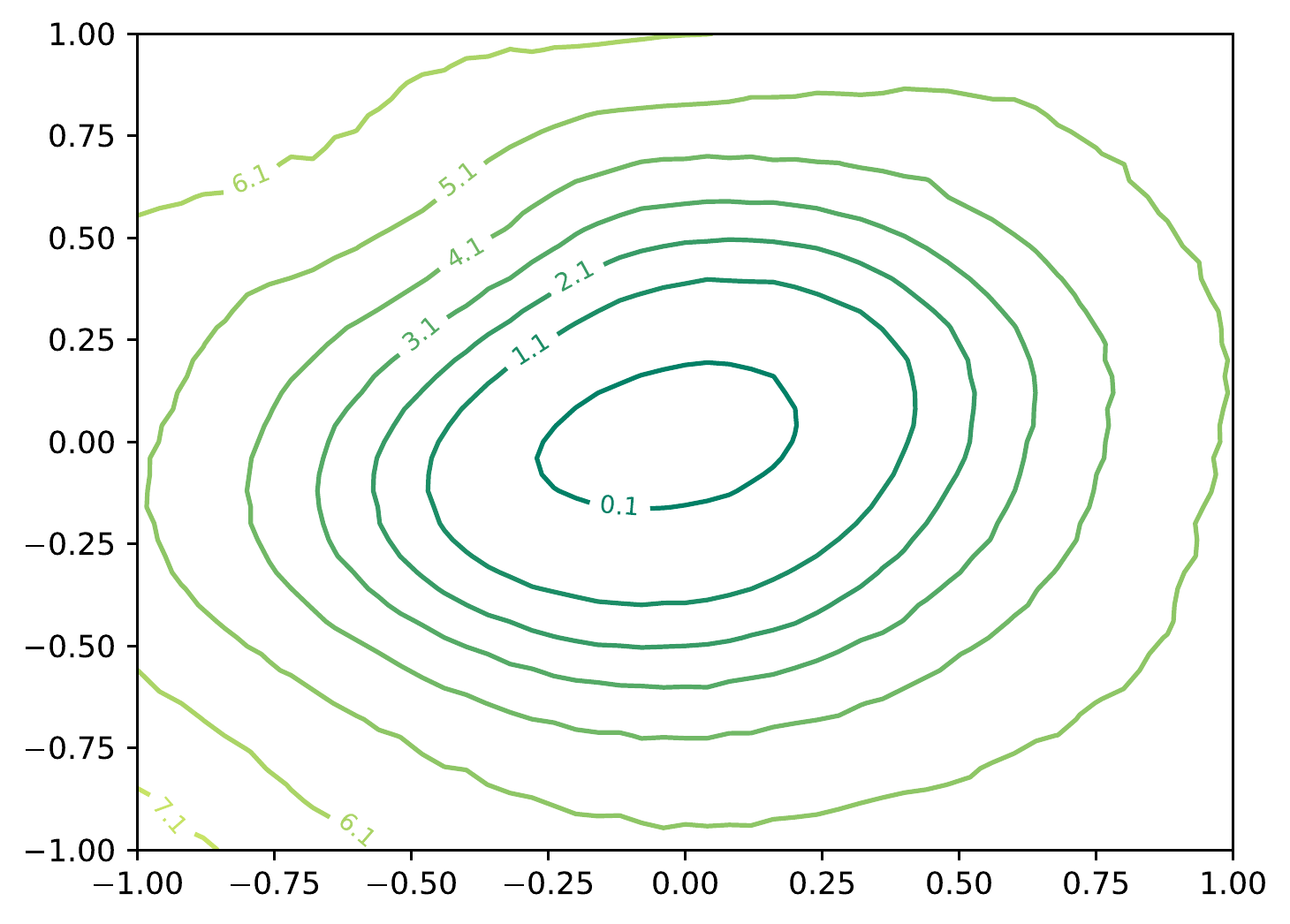}
  \label{fig:4a}
}
\subfigure[Ours, N-Caltech101]{
  \includegraphics[width=0.42\linewidth]{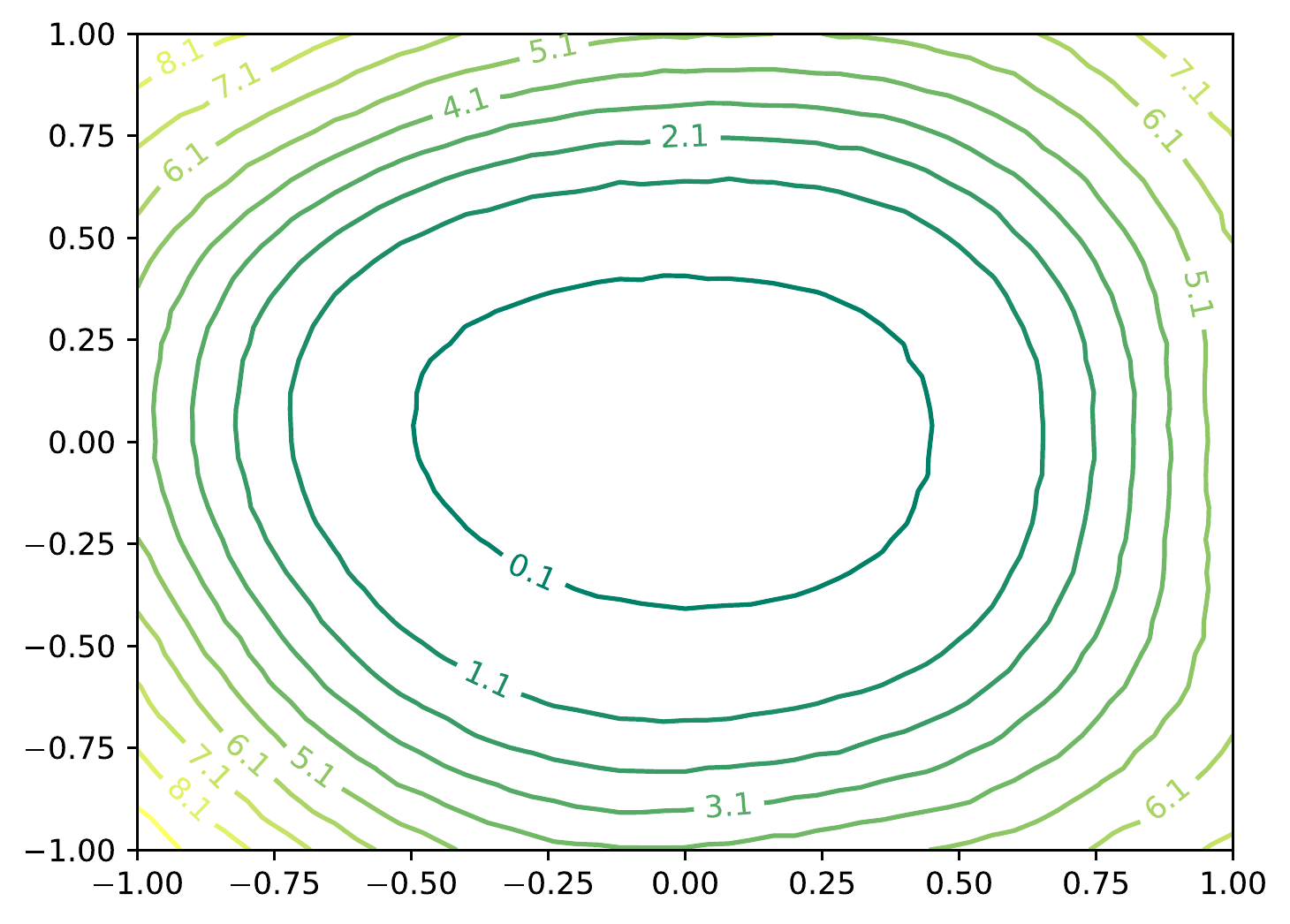}
  \label{fig:4b}
}
\subfigure[Baseline, CEP-DVS]{
  \includegraphics[width=0.42\linewidth]{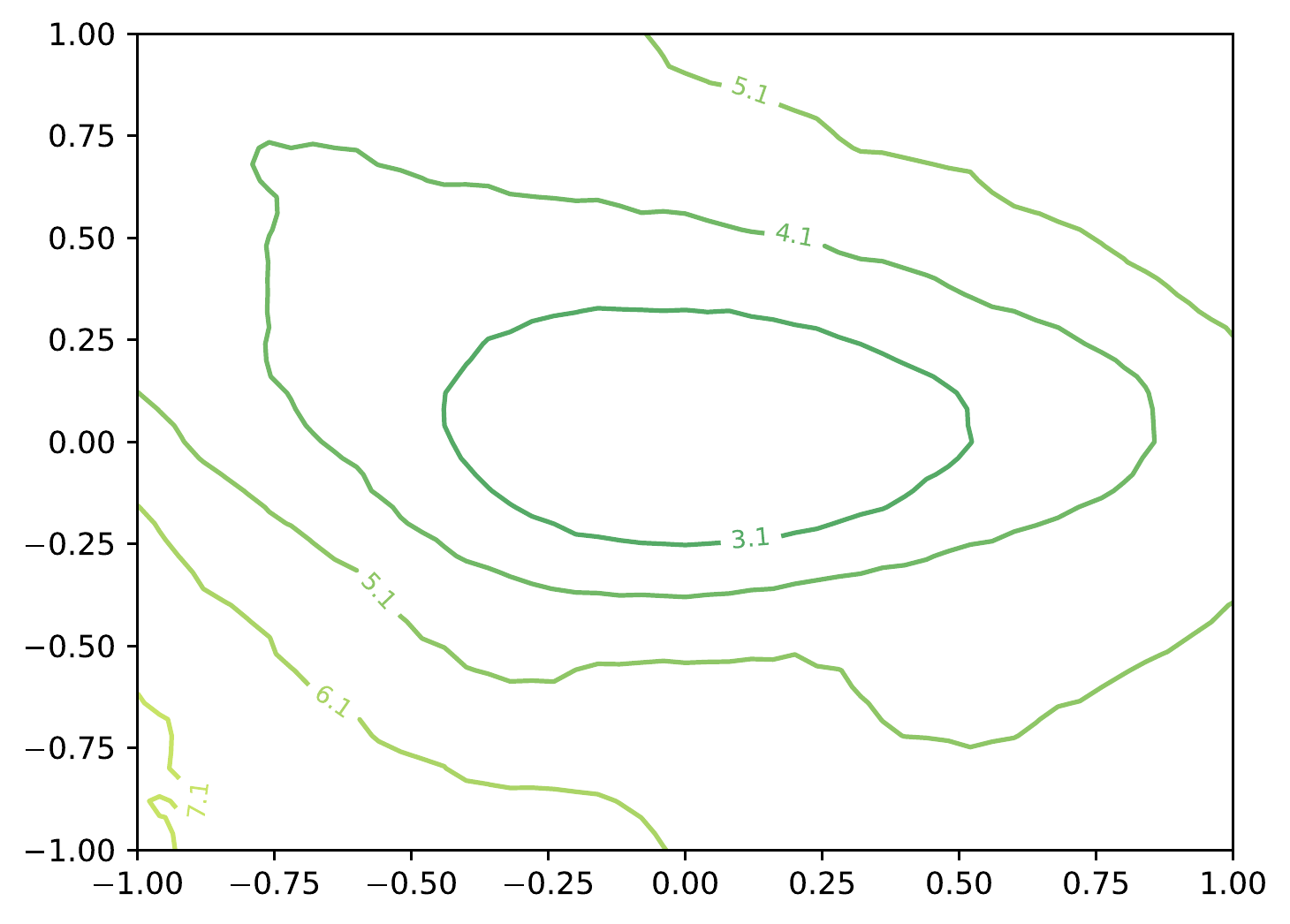}
  \label{fig:4c}
}
\subfigure[Ours, CEP-DVS]{
  \includegraphics[width=0.42\linewidth]{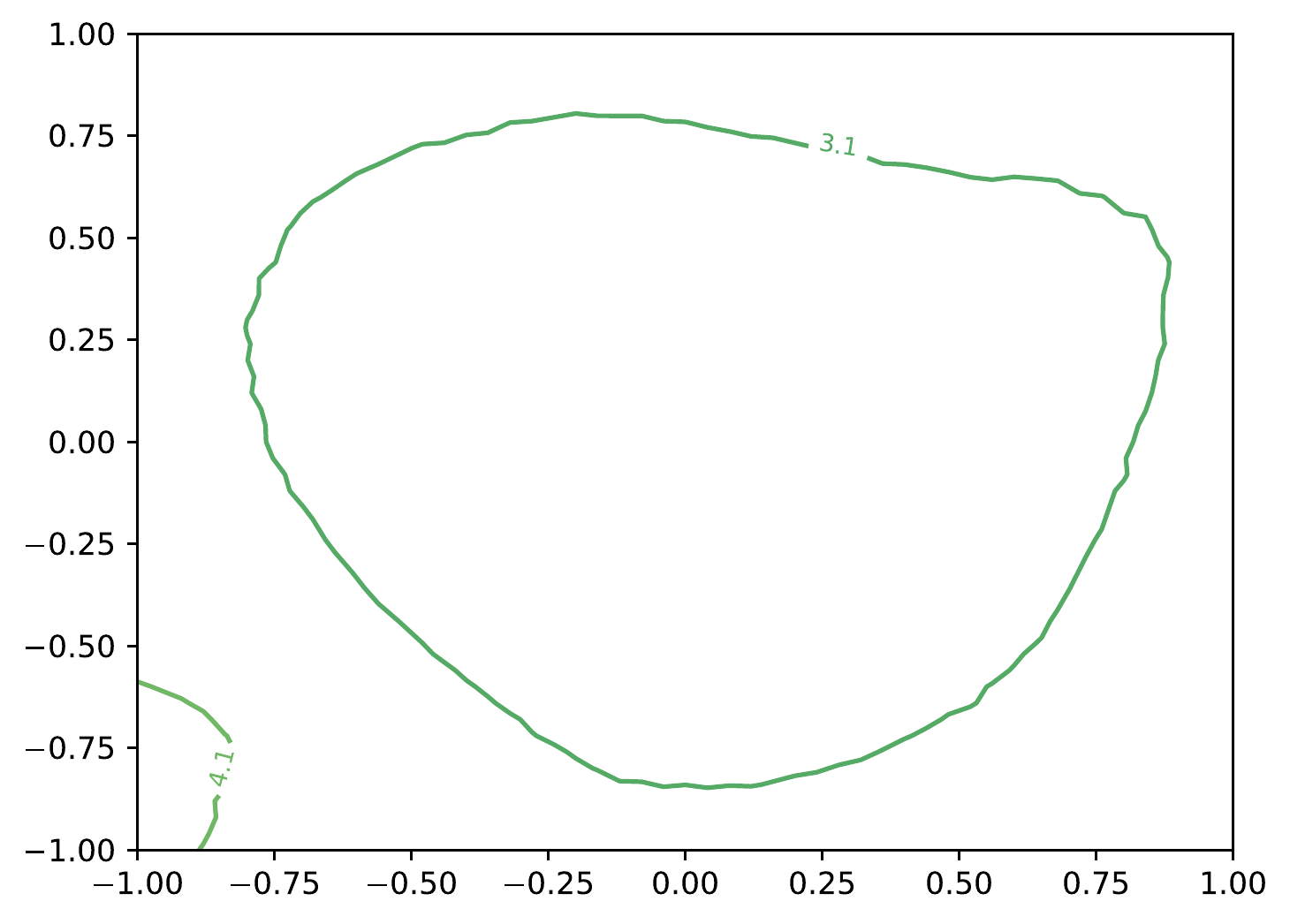}
  \label{fig:4d}
}
\caption{The loss landscape of visualization of our method and baseline on N-Caltech101 and CEP-DVS dataset.}
\label{fig4}
\end{figure}

\subsection{Analysis and Discussion}

\subsubsection{Loss Landscape.}
To verify that our method provides SNNs with more generalizability over event data, we utilize 2D loss-landscapes visualization \citep{li2018visualizing}.
To this end, we selected the optimal results of the baseline and our method to conduct experiments on N-Caltech101 and CEP-DVS respectively.
As depicted in Fig.~\ref{fig:4b} and Fig.~\ref{fig:4d}, the lowest loss area becomes flatter compared to Fig.~\ref{fig:4a} and Fig.~\ref{fig:4c}, which indicates that the SNN obtains better weights with the knowledge transfer from static images.

\subsubsection{Visual Explanations from Deep Networks.}
\begin{figure}[t]
\centering
  \includegraphics[width=0.95\linewidth]{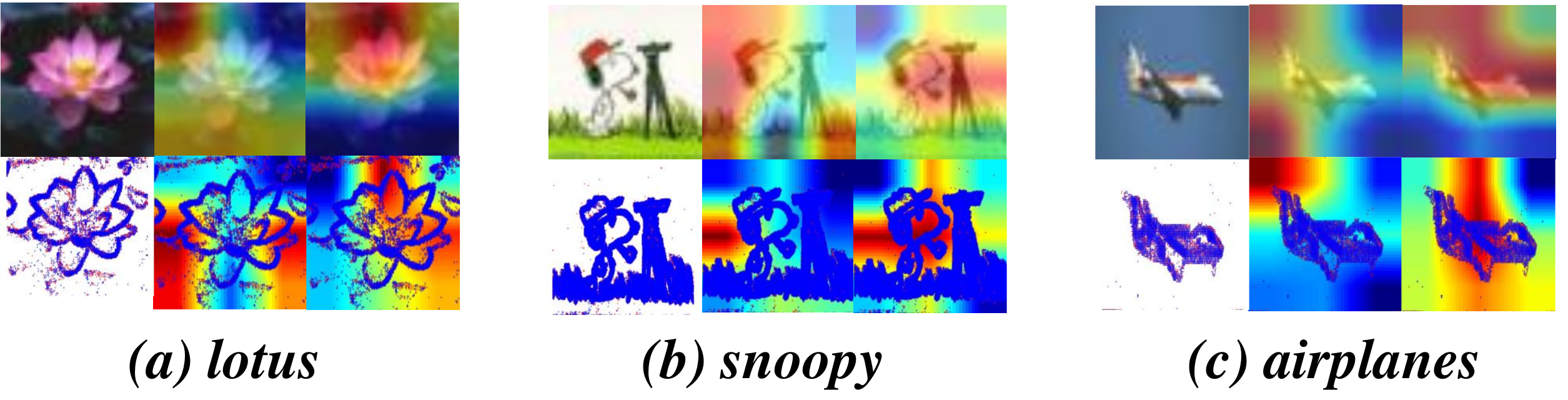}
\caption{Class Activation Mapping of Caltech101 and N-Caltech101. Three categories are selected for display, the top row under each category represents static images, and the bottom row represents event data integrated into frames. The three columns from left to right represent the results of original picture, baseline and our method, respectively.}
\label{fig1}
\end{figure}

To assess whether our method learns domain-invariant features of static images and event data, and provides helpful information for SNNs about features of event data, we employ grad-cam++~\cite{chattopadhay2018grad} visualization method. Such visualization allows us to understand which local locations of an original image contributed most significantly to the model's final classification decision. 
Ideally, static pictures and event data integrated into frames have similar object contour features when they are in the same class. This is well illustrated in Fig. \ref{fig1}, where by introducing knowledge transfer loss, for both static pictures and event data, the network pays attention to the contour features of the category. In particular, the results on event data show that our method helps SNNs to move away from the background of the event data and focus on the features of category itself.

\begin{figure}[t]
\centering
  \includegraphics[width=0.95\linewidth]{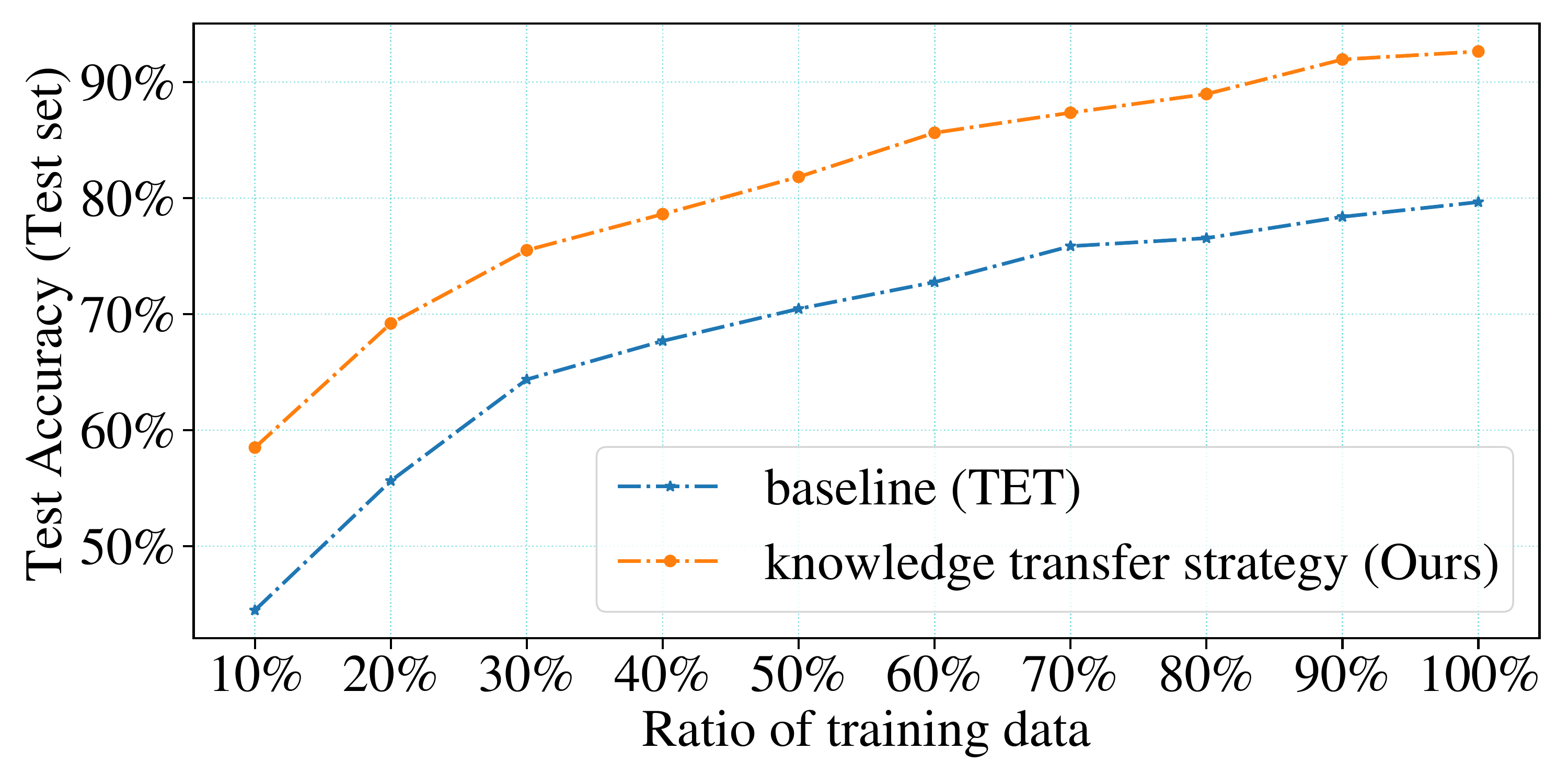}
\caption{Performance on different amounts of event data.}
\label{dvsdata}
\end{figure}

\subsubsection{Performance of Our Method on Different Amounts of Event Data.}
We conduct a detailed evaluation of our proposed approach on N-Caltech101 dataset using varying amounts of training data, as presented in Fig. \ref{dvsdata}. 
Our results show that regardless of training data amount, knowledge transfer loss results in a remarkable performance improvement. This is attributed to the knowledge transfer loss that allows the model to finely adapt to event data features, providing more generalized knowledge to the network.

\section{Conclusion}
In this paper, we explore the challenges faced by spiking neural networks when dealing with event-driven data. By using static images to assist SNN training, we improve the 
generalization ability of the network. Our proposed domain alignment loss and spatio-temporal regularization support knowledge transfer and alleviate the domain mismatch between static and event datasets. Meanwhile, we propose a sliding training strategy to bring greater stability to network training. Experiments on different event datasets show that our method achieves the best performance. In conclusion, this study not only provides new methods for training SNNs on event-driven datasets but also contributes to further development in the field of neuromorphic computing.

\section{Acknowledgments}
This work is supported by the National Natural Science Foundation of China (Grant No. 62372453).

\bibliography{aaai24}

\section{Appendix}
\externaldocument{extra}
\renewcommand{\thetable}{S\arabic{table}}
\renewcommand{\thefigure}{S\arabic{figure}}
\renewcommand{\thealgorithm}{S\arabic{algorithm}}
\renewcommand{\theequation}{S\arabic{equation}}
\setcounter{table}{0}
\setcounter{figure}{0}

\subsection{Introduction of Datasets}
An overview of the datasets used in our experiments is shown in Tab. \ref{datasets}.

\subsubsection{CIFAR100.}
The CIFAR-100 dataset \citep{krizhevsky2009learning} consists of 60,000 color images, each of size 32x32 pixels. These images are divided into 100 classes, with 600 images per class. 50,000 images are used for training and the remaining 10,000 images are used for testing.

\subsubsection{CEP-DVS.}
The cifar-event paired dataset (CEP-DVS) \cite{deng2021learning} is an event image pairing dataset that contains 10,000 samples in 20 categories. Event samples are generated by capturing motion images of the CIFAR100 dataset displayed on the monitor by an event camera.

\subsubsection{Caltech101.} 
The Caltech101 dataset \cite{fei2004learning} contains images from 101 object categories and one background category with a total of 9,145 images and approximately 40 to 800 images for each object category.

\subsubsection{N-Caltech101.} 
The N-Caltech101 dataset \cite{orchard2015converting} is a neuromorphic version of the original Caltech101 dataset. The original data is displayed on an LCD monitor while being captured by using the saccade method of camera movement. N-Caltech 101 removes the "faces" class from the original dataset to avoid confusion with "simple faces". N-Caltech 101 has 100 object classes and one background class, with a total of 8709 samples. 

\subsubsection{Omniglot.} 
The Omniglot dataset \cite{lake2015human} consists of 1,623 handwritten characters from 50 different languages, each with 20 different handwritings, and is a class of small sample handwritten character datasets. 1,200 characters are usually selected as the training set, and the remaining 423 characters as the validation set. 

\subsubsection{N-Omniglot.} 
The N-Omniglot dataset \cite{li2022n} is the first neuromorphic dataset for few-shot learning using SNNs.  The written record of strokes is reconstructed into a video of writing tracks, and then DVS is used to obtain the event records to get the neuromorphic version of Omniglot. Its number of samples is consistent with Omniglot.

\subsection{Static and Event Data Processing Methods}
\subsubsection{Processing of Static Datasets.} 
For all static datasets, we randomly select samples from the training set from the same category as the input event data for the paired input of static images and event data. 
We resize them in a bilinear interpolation manner to be consistent with the event data.  For the Omniglot dataset, the original images are all grayscale. Therefore, we replicate the single-channel images as two-channel to align with the event data dimensions.

\subsubsection{Processing of Event Datasets.} 
The N-Caltech101 and CEP-DVS datasets are uniformly resized to 48 x 48, and the training, validation and testing sets are divided according to 9:1 and 5:3:2 respectively. We use tonic~\cite{lenz_gregor_2021_5079802} package to integrate them into ten frames, six frames respectively per sample. For N-Omniglot dataset, its size and the way of dividing training and validation sets are the same as the original dataset Omniglot, i.e., 28 x 28 pixel size and 1200 class characters as training set and 423 class characters as the validation set.
The event stream is integrated into 12 frames per sample.

\begin{table}[t]
  \centering
  \begin{threeparttable}
    \resizebox{1.0\linewidth}{!}{
      \begin{tabular}{cccc}
      \toprule
        \textbf{Datasets} & \textbf{Type}  & \textbf{Categories} & \textbf{Annotated samples}\\
        \midrule
        CIFAR100 & static images & 100 & 60000 \\
        CEP-DVS & event data & 20 & 10000 \\
        Caltech101 & static images & 101 & 9145 \\
        N-Caltech101 & event data & 101 & 8709 \\
        Omniglot & static grayscale images & 1623 & 32460\\
        N-Omniglot & event data & 1623 & 32460 \\
        \bottomrule
    \end{tabular}
    }
  \end{threeparttable}
  \caption{Overview of the datasets used in our experiments.}
  \label{datasets}
\end{table}

\subsubsection{Input Dimension Alignment.}
Event data are generated based on rich localized intensity variations in continuous time; therefore, the essence of neuromorphic data describes a sequence of pixel intensity changes over time. Traditional static images use RGB color space, in which all three channels (red, green, and blue) are easily influenced by luminance, i.e., any slight change in luminance will lead to a corresponding change in these three channels. Therefore, it is not intuitive to use RGB to reflect light intensity. 
Compared with RGB color space, HSV (Hue, Saturation, and Value) color space is more suitable for dealing with light intensity changes.
Given this, we choose to convert the static image to the HSV color space to minimize the mismatch between the two types of input data, improving our model's performance and adaptability. To adapt to the dual-channel characteristics of the event data, i.e., positive and negative polarity, we replicate the value channel and then duplicate the static image in equal time steps. We feed it into the network along with the event data. We replicate directly without additional color space conversion for static image datasets with only a single grayscale channel, such as N-Omniglot.

\begin{figure}[t]
	\centering
		\includegraphics[width=1.0\linewidth]{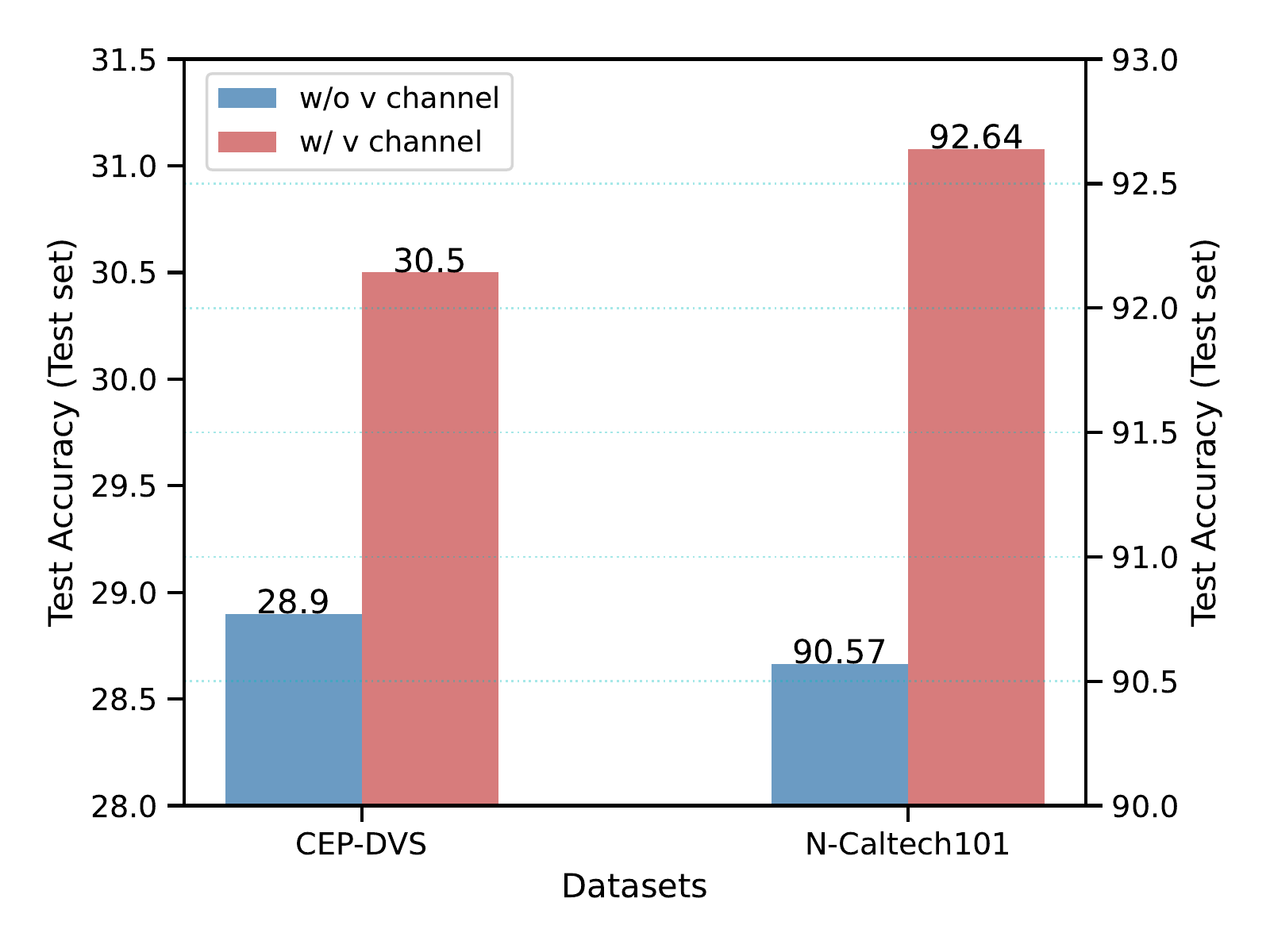}
	\caption{Influence of the value channel on classification accuracy. We use v channel to refer to value channel. }
	\label{vchannel}
\end{figure}

We conduct experiments with the VGGSNN and ResNet-18 models on the N-Caltech101 and CEP-DVS datasets respectively. 
We randomly select one of the three RGB channels to represent the without the value channel. The results are shown in Fig. \ref{vchannel}. 
It can be observed that after the conversion to HSV space, the accuracy of the model on both datasets is improved. 
This demonstrates that using the value channel of HSV color space to represent the light intensity of a static image can better match the characteristics of the event data.

\subsection{Discussion}
\subsubsection{Effect of Different Numbers of Static Images on Results.} 
\begin{figure}[t]
	\centering
		\includegraphics[width=1.0\linewidth]{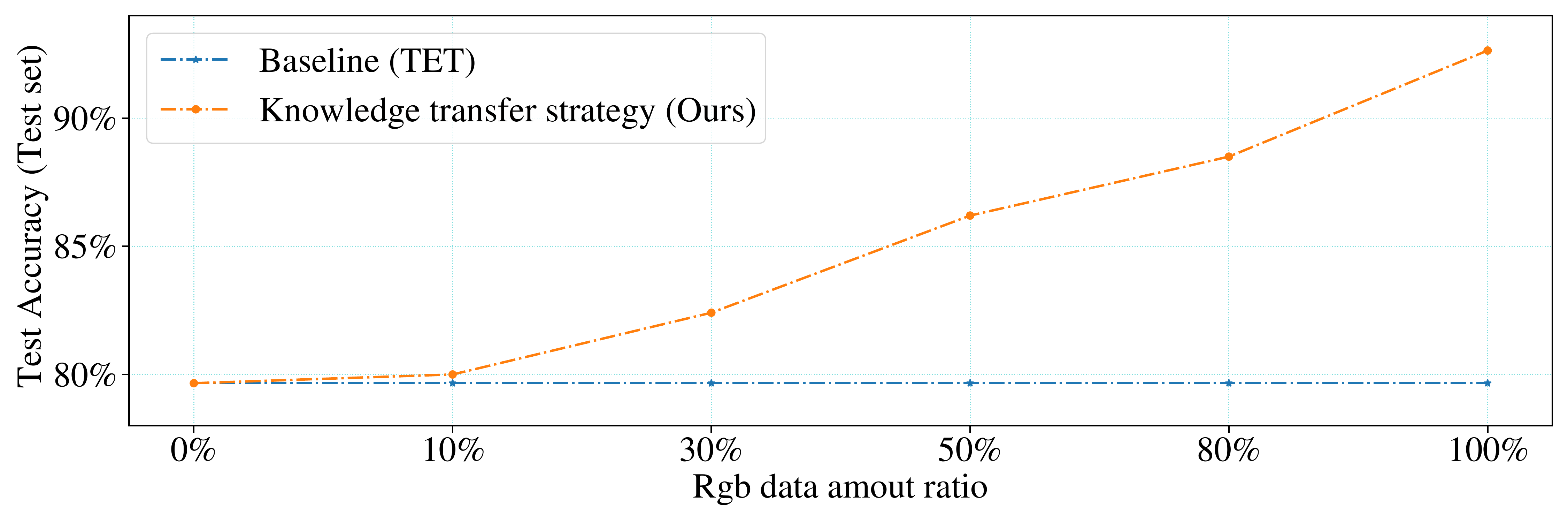}
	\caption{Performance of our method with different amounts of static data. The more static images used, the model performs better.}
	\label{rgb_enough_for_dvs}
\end{figure}

So far, we have leveraged the full amount of static images. However, a question worth exploring is: How helpful are different numbers of static images in aiding the correct classification of event data?
We experimented with VGGSNN on the N-Caltech101 dataset, and the results are shown in Fig. \ref{rgb_enough_for_dvs}. 
Considering all the event training data, 100\% static images mean the same amount of data as the event data. As can be seen from Fig. \ref{rgb_enough_for_dvs}, the more static images used, the richer the generalized features provided to the model, and thus the model performs better. 
When only 10\% of the static images are used to help the model learn the event features, the performance improvement of the model is relatively limited. However, when all the static images are utilized, the performance of the model improves significantly, with the accuracy increasing from 79.66\% to 92.64\%.

\begin{algorithm}[!t] 
  \caption{Efficient Knowledge Transfer Strategy for Spiking Neural Networks} 
  \begin{small}
  \begin{algorithmic}[1] 
      \STATE {\bfseries input:} network's parameter $\mathbf{\theta}$, SNN time step T, learning rate $\alpha$, training epoch $E$, loss $L$, manual setting epoch $e_s$, shared model $g$, classification head $h_s$, $h_t$ and a training batch sample $\mathbf{x}_s^i$ and $\mathbf{x}_t^j$ from static images and event data respectively.
      \STATE {\bfseries Ensure:} Different domain samples belong to the same category, i.e., $y_i = y_j, y \in \mathcal{Y}$
  \FOR {$e \gets 1,2, \cdots E$}
  \STATE Replace $\mathbf{x}_{s,m}^i$ with $\mathbf{x}_{t,m}^j$ in the probability of $P_{replacement}$, where $m$ is the subscript index of a batch.
  \STATE Define empty list $V_{mem,s}, V_{mem,t}, V_{mem,so}$
      \FOR {$t \gets 0, 1, \cdots T-1$}
      \STATE $V_{mem,s}[t] \gets g(\mathbf{x}_s^i) $
      \STATE $V_{mem,so}[t] \gets h_s(V_{mem,s}[t]) $
      \ENDFOR
      \STATE Reset the membrane potentials and spikes
      \FOR {$t \gets 0, 1, \cdots T$}
      \STATE $V_{mem,t}[t] \gets g(\mathbf{x}_t^i) $
      \ENDFOR
      \STATE $\mathcal{L}_{cls-s} = \mathcal{L}_{TET}(V_{mem,so}, y)$
      \STATE Compute Knowledge transfer loss $\mathcal{L}_{kt}$ as Eq .\ref{Eq15} 
      \STATE $\mathcal{L}_{all} = \lambda_{cls-s}\mathcal{L}_{cls-s}$
      \IF{ $e \leq e_s$}
      \STATE $\mathcal{L}_{all} = \mathcal{L}_{all} + \lambda_{kt}\mathcal{L}_{kt}$
      \ENDIF
  \STATE update parameter $\mathbf{\theta} = \mathbf{\theta} - \alpha \bigtriangledown_{\mathbf{\theta}}\mathcal{L}_{all}$
  \STATE Reset the membrane potentials and spikes
  \ENDFOR
  \end{algorithmic} 
\end{small}
  \label{alg:1}
\end{algorithm}

\subsubsection{Hyperparameter Settings.}
The following parameters need to be set manually in our method:
static image classification coefficient $\lambda_{cls-s}$, knowledge transfer coefficient $\lambda_{kt}$ and end of the epoch $e_s$. 
We set $\lambda_{kt}$ value to 0.5 in all cases.
For N-Caltech101 and CEP-DVS, we set $\lambda_{cls-s}$ to 1.0 and $e_s$ to maximum training epoch $e_m$ respectively.
For N-Omniglot, we we set $\lambda_{cls-s}$ to 1.0 and $e_s$ to $0.8e_m$.
Our approach can be summarized as Algorithm \ref{alg:1}.

\subsubsection{Utilization and Strengths of CKA.}
CKA is a similarity metric that utilizes a kernel approach to measure data similarity in high-dimensional feature spaces.
It is increasingly used to compare similarities in network representations, e.g. \cite{li2023uncovering}.
In our work, static and event data represent different modalities with inherent domain mismatch. In order to reduce domain distribution differences, we incorporate a distribution difference metric (actually the opposite of the similarity metric) into the loss function. 
CKA helps to reduce distribution differences between domains by measuring network similarity to align the feature spaces between static and event domains.
This alignment is well suited for capturing domain-invariant features, which optimizes the loss function and mitigates network overfitting. 
It is suitable for dealing with domain adaptation problems, including but not limited to SNN features.
With other experimental settings held constant, we replace the similarity metric with maximum mean difference (MMD) \cite{gretton2012kernel}, a widely used metric in domain adaptation. Tab. \ref{table:combined-results}(a) shows the experiment results to quantify the CKA strengths.

\begin{table}[h]
  \centering
  \begin{minipage}[t]{0.46\linewidth}
    \centering
    \begin{threeparttable}
      \resizebox{\textwidth}{!}{
        \begin{tabular}{cc}
          \toprule
          Similarity Metric & Accuracy\\
          \midrule
          - &25.70\%\\
          w/ MMD & 26.25\%\\
          w/ CKA & \textbf{30.50\%}\\
          \bottomrule
        \end{tabular}
      }
      \caption*{(a) Metric comparison}
    \end{threeparttable}
  \end{minipage}
  \hfill
  \begin{minipage}[t]{0.46\linewidth}
    \centering
    \begin{threeparttable}
      \resizebox{\textwidth}{!}{
        \begin{tabular}{cc}
          \toprule
          Training Method & Accuracy\\
          \midrule
          SEW-ResNet18 & 25.70\%\\
          Spikformer & 26.15\%\\
          ResNet18 (Ours) & \textbf{30.50\%}\\
          \bottomrule
        \end{tabular}
      }
      \caption*{(b) Method comparison}
    \end{threeparttable}
  \end{minipage}
  \caption{Experiments results on the CEP-DVS datasets.}
  \label{table:combined-results}
\end{table}

\subsubsection{Comparison with the High-Performance SNN Works.}
We have already compared the work of normal SNN training techniques in Tab. \ref{sota}, and the comparison illustrates the advantages of our method. 
Furthermore, we compare the proposed method with two efficient SNN network architectures, Deep Residual SNN \cite{fang2021deep} and Spikformer \cite{zhou2022spikformer}. We conduct the experiment with these two networks on CEP-DVS dataset. 
We implement Spikformer according to \cite{zhou2022spikformer}, using 3 transformer encoder blocks and setting the SSA head to 16, keeping other settings consistent with the paper, and the experimental results are shown in Tab. \ref{table:combined-results}(b).
As can be seen, a well-designed network structure can still be limited by the small size of the event data, and the results are not so satisfactory. This also illustrates the need to utilize static images to mitigate overfitting from another perspective.

\subsubsection{Computational Costs of Knowledge Transfer Training.}
\begin{figure}[h]
	\centering
		\includegraphics[width=1.0\linewidth]{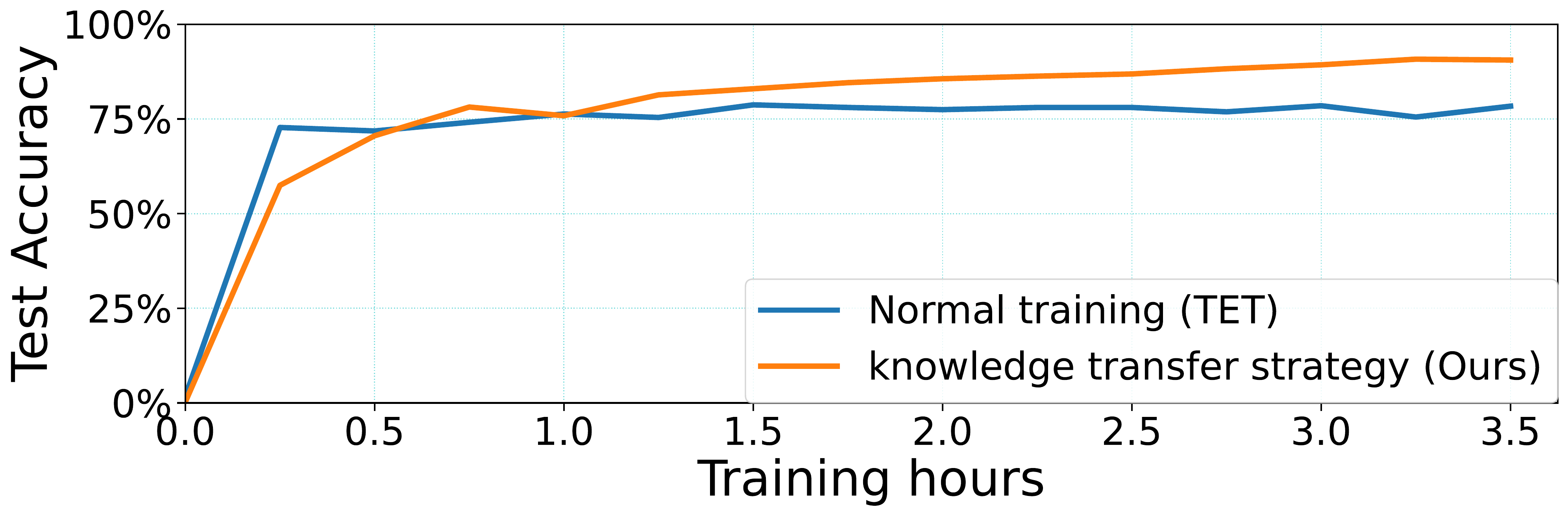}
	\caption{Variations of accuracy with the training time.}
	\label{training-time}
\end{figure}

Specifically, in order to obtain the final layer features of the static image feature extractor and align the domain distribution, our method requires one additional forward and backward propagation in each iteration. According to our experiments, this extra computation is affordable and improves the performance of the network.
In addition, our method requires fewer training iterations and, therefore, achieves better results than existing methods with the same training time, as shown in Fig. \ref{training-time}.
It is worth noting that, 
the additional computation only involves the training stage and there is no additional computational overhead in the inference stage. 

\end{document}